\pdfoutput=1

\documentclass[11pt]{article}

\usepackage{acl}

\usepackage{colortbl}

\usepackage{times}
\usepackage{latexsym}
\usepackage{graphicx}
\usepackage{booktabs}
\usepackage{stfloats}

\usepackage[T1]{fontenc}

\usepackage[utf8]{inputenc}

\usepackage{microtype}

\usepackage{inconsolata}

\usepackage[normalem]{ulem} 

\title{How Lexical is Bilingual Lexicon Induction?}

\author{
    Harsh Kohli\thanks{*Work done while at Amazon.} \\ 
    The Ohio State University \\ 
    \texttt{kohli.120@osu.edu} \And
    Helian Feng \\ 
    Amazon \\ 
    \texttt{hlfeng@amazon.com} \And
    Nicholas Dronen \\ 
    Amazon \\ 
    \texttt{ndronen@amazon.com} \AND
    Calvin McCarter \\ 
    Amazon \\ 
    \texttt{mccarter.calvin@gmail.com} \And
    Sina Moeini \\ 
    Amazon \\ 
    \texttt{smoeini@amazon.com} \And
    Ali Kebarighotbi \\ 
    Amazon \\ 
    \texttt{alikeba@amazon.com}
}

\begin{document}
\maketitle
\begin{abstract}
In contemporary machine learning approaches to bilingual lexicon induction (BLI), a model learns a mapping between the embedding spaces of a language pair. Recently, retrieve-and-rank approach to BLI has achieved state of the art results on the task. However, the problem remains challenging in low-resource settings, due to the paucity of data. The task is complicated by factors such as lexical variation across languages. We argue that the incorporation of additional lexical information into the recent retrieve-and-rank approach should improve lexicon induction. We demonstrate the efficacy of our proposed approach on XLING, improving over the previous state of the art by an average of 2\% across all language pairs.
\end{abstract}

\section{Introduction}
\label{section:Introduction}

Bilingual lexicon induction (BLI) is fundamental to many downstream NLP applications, such as machine translation \cite{qi-etal-2018-pre, duan-etal-2020-bilingual}, cross-lingual information retrieval \cite{10.1145/2766462.2767752}, document classification \cite{klementiev-etal-2012-inducing}, dependency parsing \cite{guo-etal-2015-cross, ahmad-etal-2019-difficulties}, and language acquisition and learning \cite{yuan-etal-2020-interactive}. In addition, it facilitates model sharing between high-resource and their aligned low-resource languages.

Contemporary approaches to BLI involve alignment of embeddings trained on monolingual corpora into a shared vector space. A challenge of this approach is \emph{hubness} -- the problem of high density regions in cross-lingual word embedding (CLWE) space where, in the alignment space, the embedding of a term in a source language is surrounded by a dense cluster of terms in the target language. These hub terms are difficult to align and are worthy of investigation. The recent cross-domain similarity local scaling (CSLS) addresses this by normalizing distances by the average distance of each term's embedding to its nearest neighbors \cite{conneau2017word}. While it would be desirable to take advantage of CSLS in a state-of-the-art BLI model such as BLICEr \cite{li-etal-2022-improving-bilingual}, computing nearest neighbors is prohibitively expensive. While performance is better due to a pairwise cross-attention mechanism, this affects our ability to perform an approximate nearest neighbour lookup.

\begin{figure*}
  \includegraphics[width=\textwidth]{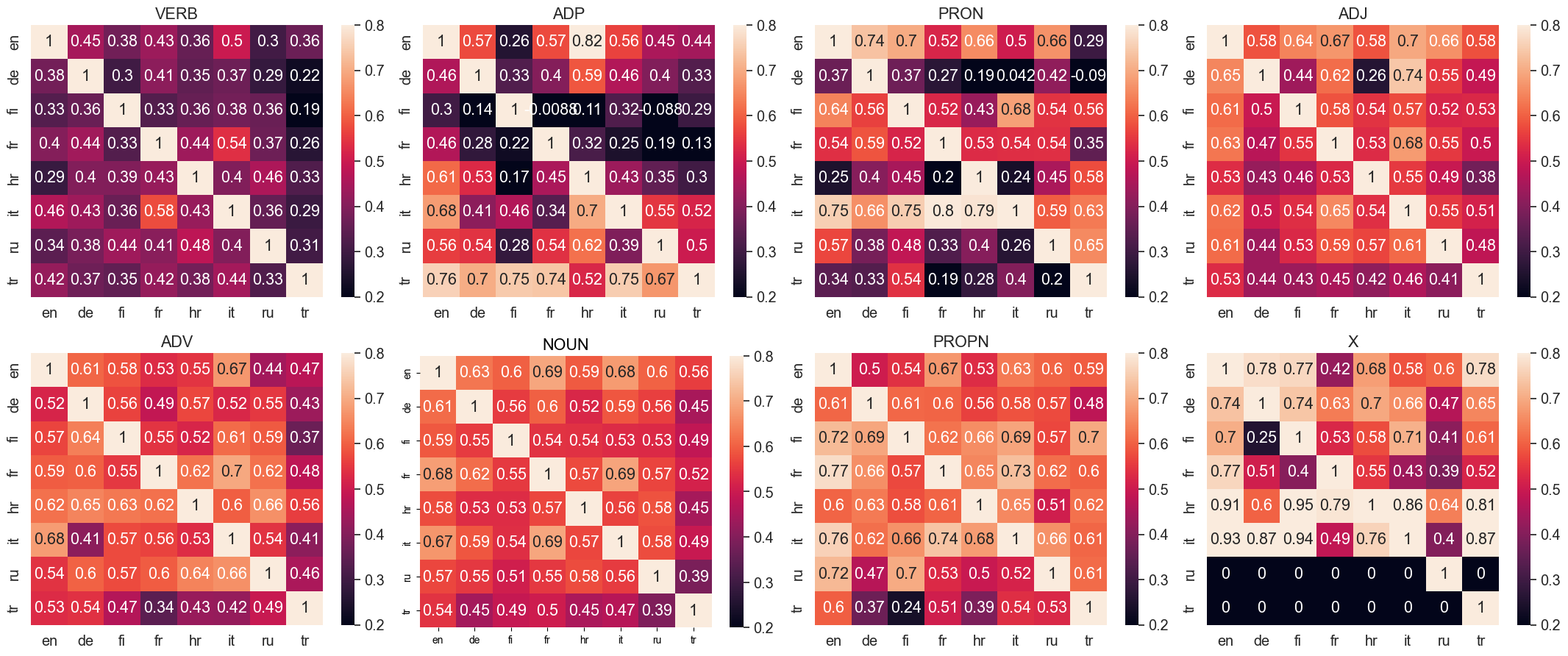}
    \caption{\textbf{Spearman's Rank correlation of term frequencies derived from Common Crawl and Wikipedia.} The plot visualizes the Spearman's Rank correlation of term frequencies between each of the source (by row) and target (by column) language pairs in the 5k vocabularies in  the XLING corpus derived from Common Crawl and Wikipedia. We calculate and plot the correlation heatmap separately by part of speech in each of the sub-figures. Cells containing a 0 have an insufficient (<10) number of terms in the source language for a particular part of speech.}
  \label{fig:pos}
\end{figure*}

We propose instead to address the hubness problem by including simple lexical features. We start with the observation that the lexical similarity of a pair of languages tends to be indicated by a relatively high rank correlation of term frequency, particularly for certain parts of speech (POS). Figure~\ref{fig:pos} shows, by part of speech, the Spearman's rank correlation of corresponding terms in the 5k vocabularies in the XLING corpus \cite{glavas-etal-2019-properly}. As we can see from the plot, 1) all language pairs have a positive rank correlation for term freq across all part of speech; 2) the correlation varies by part of speech, for example, the term freq correlation is highest for proper nouns (\texttt{PROPN}) and and nouns and the least so for verbs. This suggests that including term frequency and part of speech as features to the model can improve alignment of terms in high-density regions of the embedding space. Indeed, our approach improves the state of the art by 2.75\% and 1.2\% on the semi-supervised and supervised splits, respectively, of the XLING benchmark. An additional benefit of our approach is that it does not incur the computational overhead of the more complex CSLS for the pairwise approach.

\section{Related Work}

Popular methods for BLI using embedding alignment include Canonical Correlation Analysis (CCA) \cite{Faruqui2014ImprovingVS}, minimizing euclidean distance \cite{mikolov2013exploiting, artetxe-etal-2016-learning}, and methods relying on Procrustes alignment of word embedding spaces \cite{conneau2017word, Grave2018UnsupervisedAO, alvarez-melis-jaakkola-2018-gromov}. A ranking-based objective was introduced by \citet{joulin-etal-2018-loss} which aims to minimize the CSLS metric. The VecMap approach \cite{Artetxe2018GeneralizingAI} achieves impressive results through a series of pre-processing (length normalization, mean centering, ZCA whitening) and post-processing (dimensionality reduction, cross-correlation weighting) steps. 

Contrary to most other NLP tasks, static word embeddings such as fastText \cite{bojanowski2017enriching} continue to be the de facto choice of base embedding on BLI due to their strong performance. Promising results have been obtained by introducing an orthogonality constraint on the linear projection matrix \cite{Xing2015NormalizedWE}. However, non-linear projection techniques such as LNMap \cite{mohiuddin-etal-2020-lnmap} have demonstrated better results on many language pairs. Over time, several efforts  have been made to explore the limitations of using a linear projection on the BLI task \cite{10.1162/coli_r_00372, ormazabal-etal-2019-analyzing} and explain degradation due to linguistic and domain divergences. The robustness of the isomorphic assumption is empirically challenged by \citet{sogaard-etal-2018-limitations}. \citet{DBLP:journals/corr/abs-2004-01079} proposed a metric $S_{pae}$ that gauges preservation of analogy encodings, the goal of which is to test the applicability of the linearity constraint between languages.

Methods for BLI can be classified into unsupervised, semi-supervised, and supervised approaches. While purely unsupervised methods \cite{conneau2017word, Grave2018UnsupervisedAO} have yielded impressive results on many language pairs, minimal supervision through a small seed dictionary has improved performance considerably, especially on low-resource languages. Supervised and semi-supervised approaches typically assume a dictionary of 5k and 1k word correspondences respectively for their training. In the semi-supervised setting, high-confidence alignments at each step are iteratively added as anchor points for subsequent training runs. Results on semi-supervised BLI have shown to improve by adopting a classification-based approach to iteratively refine and augment the seed translation dictionary \cite{karan-etal-2020-classification}. This also allows for including arbitrary features such as term frequency and sub-word information.

Recent semi-supervised and supervised approaches include ContrastiveBLI \cite{li-etal-2022-improving} and BLICEr. These achieve state-of-the-art results and serve as strong baselines for our work. ContrastiveBLI uses a bi-encoder with hard negative sampling and contrastive learning. Two configurations for the bi-encoder are used:

\begin{itemize}
    \item \textbf{C1}: Fine-tuned bi-encoder on static fastText.
    \item \textbf{C2}: Fine-tuned bi-encoder on multi-lingual BERT \cite{DBLP:journals/corr/abs-1810-04805}.
\end{itemize}

C2 involves an additional step of a Procrustes mapping from C1 (300-dim) to the fine-tuned BERT (768-dim) embedding. The final embeddings are then a linear combination of the projected C1 and BERT representation. 

BLICEr further improves performance through a re-ranking step using a fine-tuned cross-encoder based on \texttt{xlm-roberta-large} \cite{DBLP:journals/corr/abs-1911-02116}. Instead of a simple binary classification over sampled hard negatives, a score polarization technique is described which increases or decreases CSLS scores on a base CLWE embedding (C1 or C2) based on the assigned label. The model is then trained to predict this score. Results in BLICEr include an additional step of linearly combining the cross-encoder score with CSLS of the base embedding for each candidate. We frequently refer to C1, C2, and BLICEr in subsequent sections.

Another recent work for BLI is the BiText mining approach by Haoyue Shi et al \cite{DBLP:journals/corr/abs-2101-00148}. The authors demonstrate that a combination of unsupervised bitext mining and unsupervised word alignment leads to improved performance on the BUCC 2020 \cite{rapp-etal-2020-overview} dataset. Since current methods such as ContrastiveBLI and BLICEr do not evalute on BUCC 2020, we cannot directly compare their relative performance.

\section{Method}


\subsection{Retriever}

We use the fastText-based C1 model described previously to retrieve top candidates for our reranker. C2, which leverages both fastText and multilingual-BERT, achieves better results both as a standalone BLI system as well as when used as a retriever in BLICEr. However, for simplicity, we only use the static fastText-based C1 model in our system and note that further improvement might be had from utilizing C2 as the retriever. For the supervised and semi-supervised systems, we utilize the C1 model trained on 5k and 1k data respectively. Consistent with recent work in BLI, we use the CSLS metric to score the nearest neighbors.

\subsection{Base Reranker}

Our ranking approach closely follows BLICEr in several respects. We score each source-target candidate pair using \texttt{xlm-roberta-large}\footnote{Available via \href{https://huggingface.co/xlm-roberta-large}{https://huggingface.co/xlm-roberta-large}.}. The pairs are formatted -- e.g., for English \textit{apple} and French \textit{pomme} -- as \texttt{apple (english), pomme (français)!}. Also like BLICEr, we mine twenty hard negatives for each positive example to train the cross encoder for a binary classification objective.

While BLICEr demonstrated improvement in the supervised setting through score polarization, we maintain the simple binary objective in all our experiments. In the semi-supervised set, we use additional 4k high-confidence pairs from C1 to augment the initial 1k seed dictionary. The model is fine tuned for one epoch on each language pair.

\subsection{Learning to Rank with XGBoost}

We model our additional lexical features through a learning to rank \cite{10.1145/1273496.1273513} objective using XGBoost \cite{Chen:2016:XST:2939672.2939785}. Our method, Lexical-Feature Boosted BLI (\textbf{LFBB}), takes as input the features corresponding to the source word and all of its candidates. The following features are used in our LFBB model:

\textbf{POS features:} Source and candidate POS (categorical), and a binary label indicating a POS match. We use various Spacy \citep{Honnibal_spaCy_Industrial-strength_Natural_2020} compatible taggers to derive POS tags in each language. While inaccuracies may occur due to tagging errors, particularly in the low resource languages, our results indicate that including these POS features has a net positive effect on model performance.

\textbf{Frequency features:} Frequency ranks for the source and candidate described in Section \ref{section:Introduction}. In addition, we use the log-normalized raw frequency of source and candidate using wordfreq \cite{robyn_speer_2022_7199437} which is derived from 8 different monolingual text corpora. We separately include the difference in frequency of the source-candidate pair.

\textbf{Retriever \& Reranker features:} Raw logits returned from the base reranker (XLM-R) and CSLS score from the retriever (C1) for each pair.

Due to polysemy and synonymy, a group of candidates can consist of multiple positives as a result of synonymy in the target language. The listwise learning objective effectively shepherds our model into making better choices by taking into account relative candidate scores, their frequency alignment with the source and the part-of-speech information.

\section{Training Parameters}

For each source word in our seed dictionary (including the augmented words in the semi-supervised case), we retrieve the top 50 candidates from the target language using C1 embeddings. Negative pairs for our base reranker are sampled from these candidates. The reranker is trained for 1 epoch using a batch size of 8 and a small learning rate ($10^{-5}$). The LFBB reranker is trained using gradient-boosted decision trees with 200 estimators, maximum depth of 3, and learning rate of 0.1. All of the top 50 retrieved candidates are used in training and inference using the rank:map (mean average precision) learning objective.

\section{Results}

\begin{table*}[t]
    \scriptsize
    \centering
    \begin{tabular}{lccccccccccccc}
    \toprule
         &\textbf{en-de}&  \textbf{en-fi}&  \textbf{en-fr}&  \textbf{en-hr}&  \textbf{en-it}&  \textbf{en-ru}&  \textbf{en-tr}&  \textbf{de-*}& \textbf{fi-*}& \textbf{hr-*}& \textbf{it-*}&\textbf{ru-*} &\textbf{tr-*}\\
         \midrule
         \multicolumn{1}{c}{1k} & & & \\
         \cmidrule{1-1}
             C1&50.4&  42.15&  61.65&  35.65&  59.60&  42.50&  38.15&  41.89& 35.81& 40.26& 65.63& 48.61&32.06\\  
             C2&50.85&  45&  62.5&  42.35&  61.05&  46.05&  41.05&  44.75& 39.39& 44.68& 66.77& 50.26&35.57\\  
             RCSLS+BLICEr&56.5&  45.9&  63.65&  41.1&  64.45&  52.25&  40.2&  -& -& -& -& -&-\\
             C1+BLICEr& 52.5& 50.95& 64.4& 49.3& 65.05& 50.8& 46.55& -& -& -& -& -&-\\
             C2(C1)+BLICEr& 51.05& 50.15& 63& 50.9& 62.85& 52.7& 46.35& -& -& -& -& -&-\\
             XLM-R (Ours)& 46.45& 49.3& 58.75& 47.7& 57.9& 51.8& 40.7& 40.11& 38.89& 44.92& 58.26& 44.47&33.76\\
LFBB(XLM+CSLS)& 52.75& 50.7& 63.15& 49& 62.55& 52.75& 45.4& 45.24& 43.18& 48.79& 63.57& 51.22&38.22\\
             LFBB+Freq& 56.9& 53& 67.3& 50.7& 66.25& 54.7& 47.4& 45.74& 44.72& 50.55& 66.12& 52.88&39.32\\
             LFBB+POS+Freq& 58.2& 53.15& 67.3& 50.75& 66.3& 54.75& 47.74& 46.51& 44.98& 50.44& 67.22& 53.04&39.25\\
             LFBB+POS+Freq+C1& \textbf{58.9}& \textbf{53.45}& \textbf{68.5}& \textbf{51.9}& \textbf{67.8}& \textbf{56.45}& \textbf{49.2}& \textbf{48.88}& \textbf{46.47}& \textbf{51.54}& \textbf{68.61}& \textbf{54.79}&\textbf{40.97}\\
         \midrule
         \multicolumn{1}{c}{5k} & & & \\
         \cmidrule{1-1}
C1&54.9&  44.6&  65.05&  40.7&  63.45&  49.15&  41.35&  44.21& 39.21& 43.18& 66.51& 50.1&35.38\\ 
             C2& 57.75& 47.17& 67.2& 47.2& 65.6& 50.5& 44.74& 47.17& 42.71& 48.22& 67.86& 52.33&38.66\\ 
             CSLS + BLICEr& 64& 53.6& 71.75& 53.15& 70.5& 60.45& 50.35& -& -& -& -& -&-\\ 
             C1+BLICEr& 62.75& 54.25& 70.75& 55.4& 70.05& 59.25& 51.05& -& -& -& -& -&-\\
             C2(C1)+BLICEr& 63.45& 55.95& 70.90& \textbf{57.55}& 70.25& 60.4& 52.85& -& -& -& -& -&-\\
             XLM-R (Ours)& 52.8& 49.45& 59& 49.45& 60.04& 54.5& 41.75& 41.96& 39.04& 43.26& 54.6& 45&30.91\\
             LFBB(XLM+CSLS)& 61.2& 54.2& 68.2& 54.1& 69.2& 57.6& 50.15& 49.47& 46.46& 50.54& 66.45& 53.86&41.03\\
             LFBB+Freq& 64.75& 56.05& 71.45& 55.9& 71.6& 59.95& 51.55& 50.42& 48.6& 52.26& 68.6& 54.97&42.62\\
             LFBB+POS+Freq& 64.75& 57& 72.4& 56.65& 72.6& 61.05& 52.35& 51.57& 48.6& 53.1& 69.96& 56.08&42.57\\ 
             LFBB+POS+Freq+C1& \textbf{65.85}& \textbf{57.65}& \textbf{72.65}& 57.05& \textbf{72.85}& \textbf{61.3}& \textbf{53.3}& \textbf{52.06}& \textbf{49.29}& \textbf{53.93}& \textbf{70.94}& \textbf{56.81}&\textbf{43.22}\\ 
    \bottomrule
    \end{tabular}
    \caption{Accuracy (P@1x100) on XLING with 5k (supervised) and 1k (semi-supervised) data.}
    \label{tab:results_table}
\end{table*}

We conduct our experiments on XLING which is a widely-used standard for BLI comprising 28 language pairs from 8 different languages. We choose XLING for its good mix of languages of differing typological similarities compared to previous benchmarks \cite{conneau2017word}. The results from our modelling are presented in Table~\ref{tab:results_table}. To directly compare with baselines, we use the P@1 metric which is the probability of selected the correct correspondence in our top prediction. We benchmark our results against BLICEr used in conjunction with different retrieval backbones - RCSLS \cite{joulin-etal-2018-loss}, C1, and C2. BLICEr only reports results on en-* XLING pairs, but we also report mean unidirectional accuracy of all other language pairs and compare results with C1 and C2 which are the best reported results on those pairs. LFBB-* rows use as input the raw logits from our own version of the fine-tuned XLM-R cross-encoder model. While this model is competitive with other baselines in the semi-supervised task, its standalone results are less impressive on the fully supervised set. This difference may be attributed to a more sophisticated sampling strategy and score polarization in BLICEr. We report results with a simple LFBB model using just the XLM-R logits and CSLS score, and also incremental changes from incorporating each of the features.


\begin{figure*}
  \includegraphics[width=\textwidth]{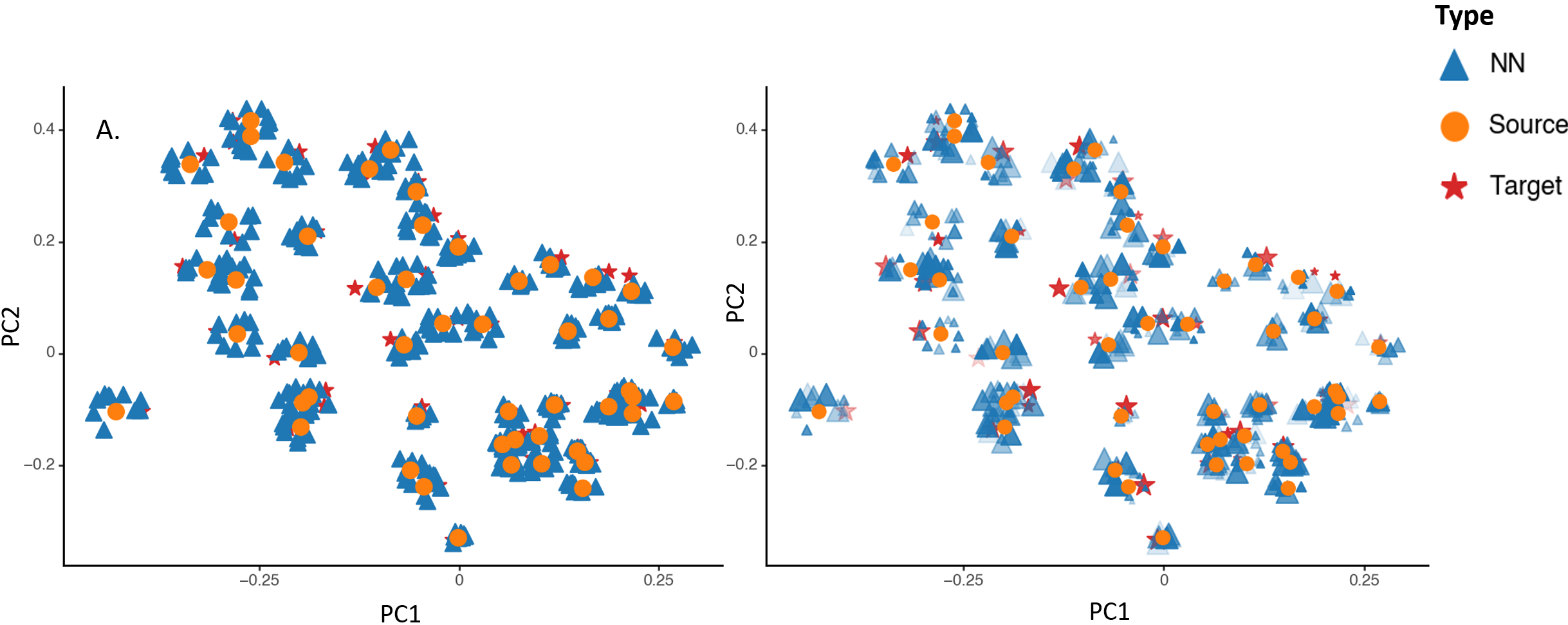}
    \caption{ \textbf{Principle Component Analysis (PCA) of source, target word and Nearest Neighbours (NN) of source word in the embedding space.} The source word, target word, and nearest neighbours are distinguished by the shape and color of points (as shown in the legend). In the left panel, we keep the size and transparency of all points the same. In the right panel, the size of the dots are scaled with the likelihood of matching POS between source and target; and the alpha (transparency) of the dots as the normalised frequency difference of source-candidate pair.}
  \label{fig:improvement}
\end{figure*}

While we used XLM-R base, our final model still outperforms BLICEr in all en-* pairs on the 1k set, and 6 out of 7 pairs on the 5k set. Due to a more competitive cross-encoder baseline, the difference is more pronounced on the 1k set. We observe results from incorporating only the frequency-based features, as well as both POS and term frequency in our reranker. Part-of-speech information improves model accuracy in most cases, albeit marginally, however, best results are obtained when both features are used in conjunction. We further analyzed improvements on a per-POS basis and discovered the largest gains for nouns (7.3\%) from amongst the most frequent POS types. This is consistent with our expectations in Figure~\ref{fig:pos}. Finally BLICEr reports results from using a linear combination of similarity scores using the cross-encoder as well as the CLWE backbone. For a more direct comparison, we do the same with our CLWE retriever (C1) which helps improve model performance across the board. Our approach yields improved results even in the absence of this additional step.

In Figure \ref{fig:improvement}, we visualize a random sample of 50 baseline error cases in the en-de test set corrected by our LFBB model with (right) and without (left) the additional lexical features. Through the re-scaling of size of points with the probability of the POS matching, and transparency by frequency difference between source and candidate pairs, we observe how these features help the target stand out better in the right panel. This illustrates how our method tackles the hubness issue. While it is hard to disambiguate between close candidates in the embedding space, the LFBB model is able to turn to external cues in the form of these lexical features to help it make better predictions.


\begin{figure}[h]
\includegraphics[width=\columnwidth, height=6cm]{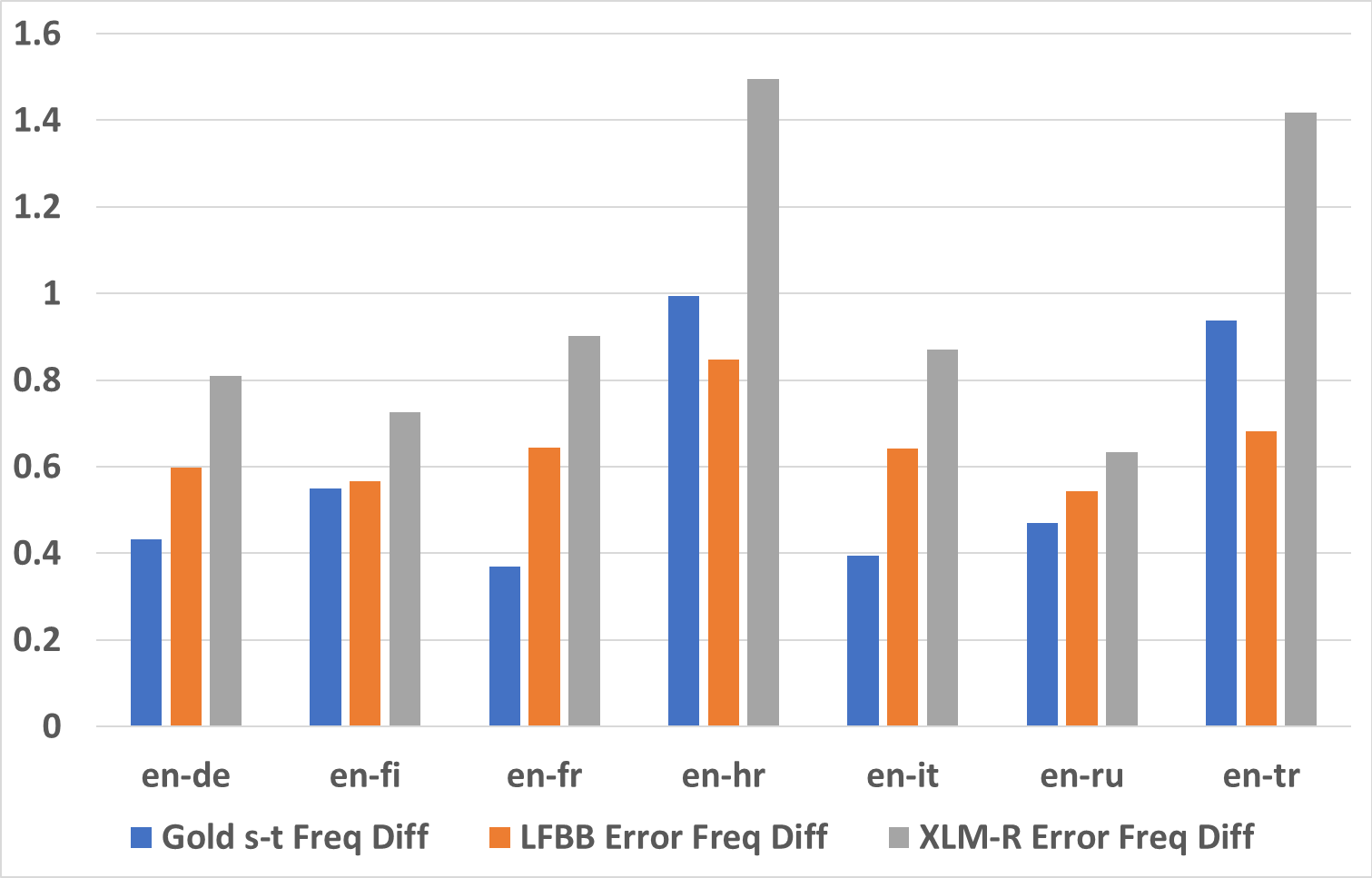}
\caption{\textbf{Mean absolute difference of term frequency.} The figure plots the mean absolute difference of term frequency between the source and target word for the ground truth, LFBB (ours), and XLM-R.}
\label{fig:x}
\end{figure}

To better hone in on how the use of these lexical features affects a model, we do a post-hoc error analysis of our model on the test set using mean absolute difference of term frequency. Figure~\ref{fig:x} shows the frequency difference in en-* pairs for the gold set and all error cases of XLM-R and LFBB. XLM-R consistently has higher frequency difference between source-predicted pairs. Conversely, predictions from the LFBB model have a frequency deviation that is more in-line with the gold distribution illustrating the models' higher proclivity to choose candidates with similar frequency. This implies that our model tends to make predictions that are closer, frequency-wise, with the source word; in-line with the labelled data.

\section{Conclusion}

Approaches to BLI have evolved to include full transformer based reranking methods. However, results on recent benchmarks indicate considerable scope of improvement still, particularly for low-resource or lexically dissimilar language pairs. While embeddings afford a rich semantic representation of individual words, we look towards supplementary features derived from individual monolingual corpora. Owing to the \textit{hubness} issue we often retrieve many close candidates highlighting the need for better reranking and additional tools to deduce the correct correspondence. Our simple-yet-effective strategy of modeling lexical features using a ranking objective yields significant improvement over baselines. We are able to quantify their impact and demonstrate the efficacy of our approach across a wide array of language pairs. We hope this work inspires further research into both the acquisition and modelling of such features to further advance state of the art on bilingual lexicon induction.


\section{Limitations}

Our proposed approach uses a relatively simple learning-to-rank approach with XGBoost. This might be less effective at capturing complex, non-linear interactions between our features (POS types, term frequency, score from upstream models) than more sophisticated approaches such as Neural Networks. Also, as noted previously, we do not use the SOTA bi-encoder based model (C2) during our retrieval step due to compute and time constraints of training BERT-based bi-encoders for each individual language pair. Similarly we do not use scores from the SOTA cross-encoder, BLICEr, as input to the LFBB model. For these reasons, our approach might not fully exploit the extent of improvements made possible by incorporating such lexical features in the BLI task.

Another limitation of our work stems from ambiguity in the evaluation set of our benchmark dataset - XLING. Samples in XLING are constructed using word tuples derived from Google Translate. This approach does not account for issues arising due to polysemy and synonymy. The test set consists of a single target correspondence for each source word when, in practice, multiple correspondences might exist. Thus, a performance measure of any model evaluated on this test set, while indicative, does not fully reflect its efficacy on this task.

\bibliography{anthology,custom}


\begin{figure*}[t!] 
\noindent 
\begin{minipage}{0.5\textwidth} 
  \centering
    \includegraphics[width=\linewidth]{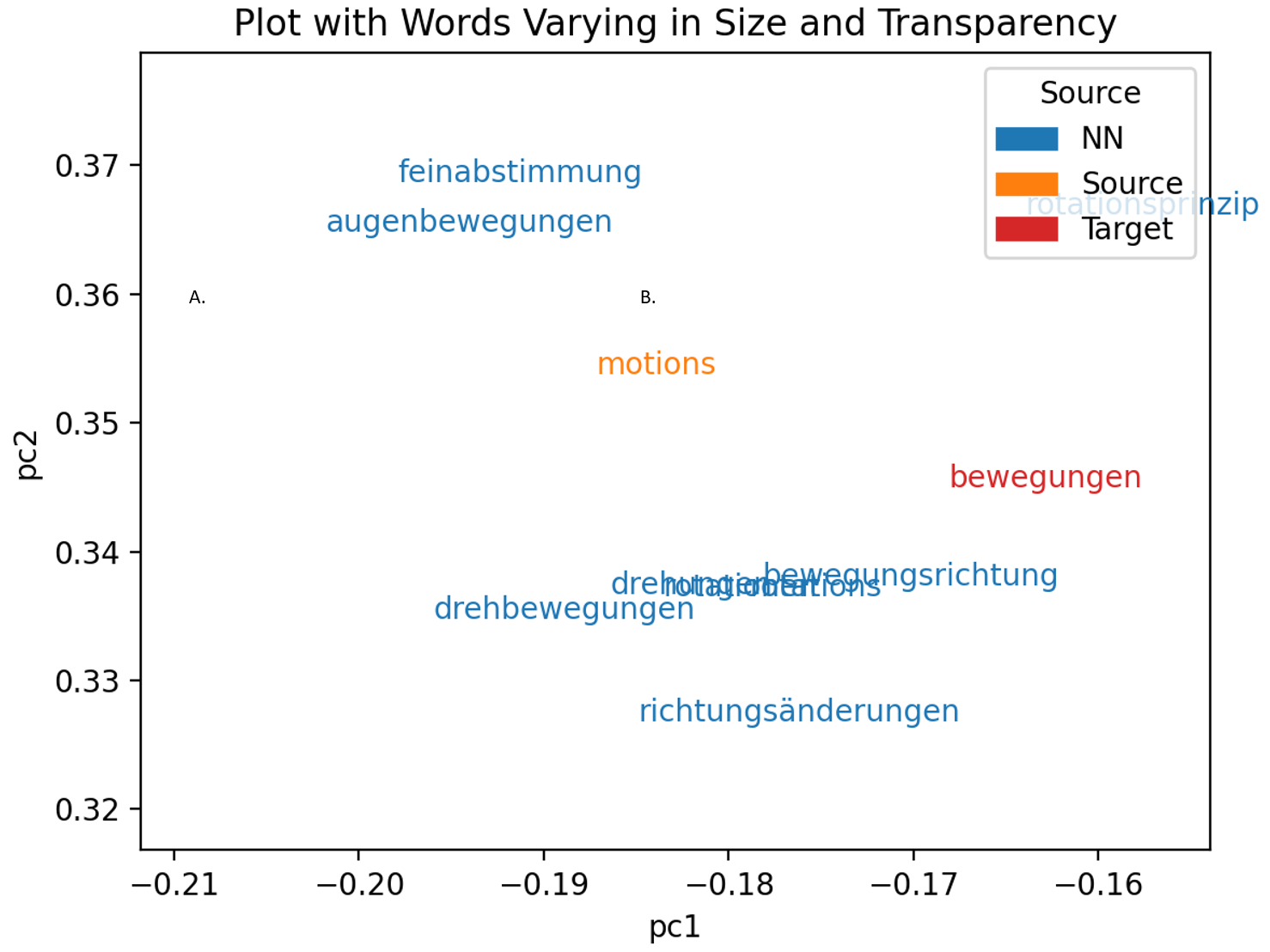}
\end{minipage}%
\hfill 
\begin{minipage}{0.5\textwidth} 
  \centering
    \includegraphics[width=\linewidth]{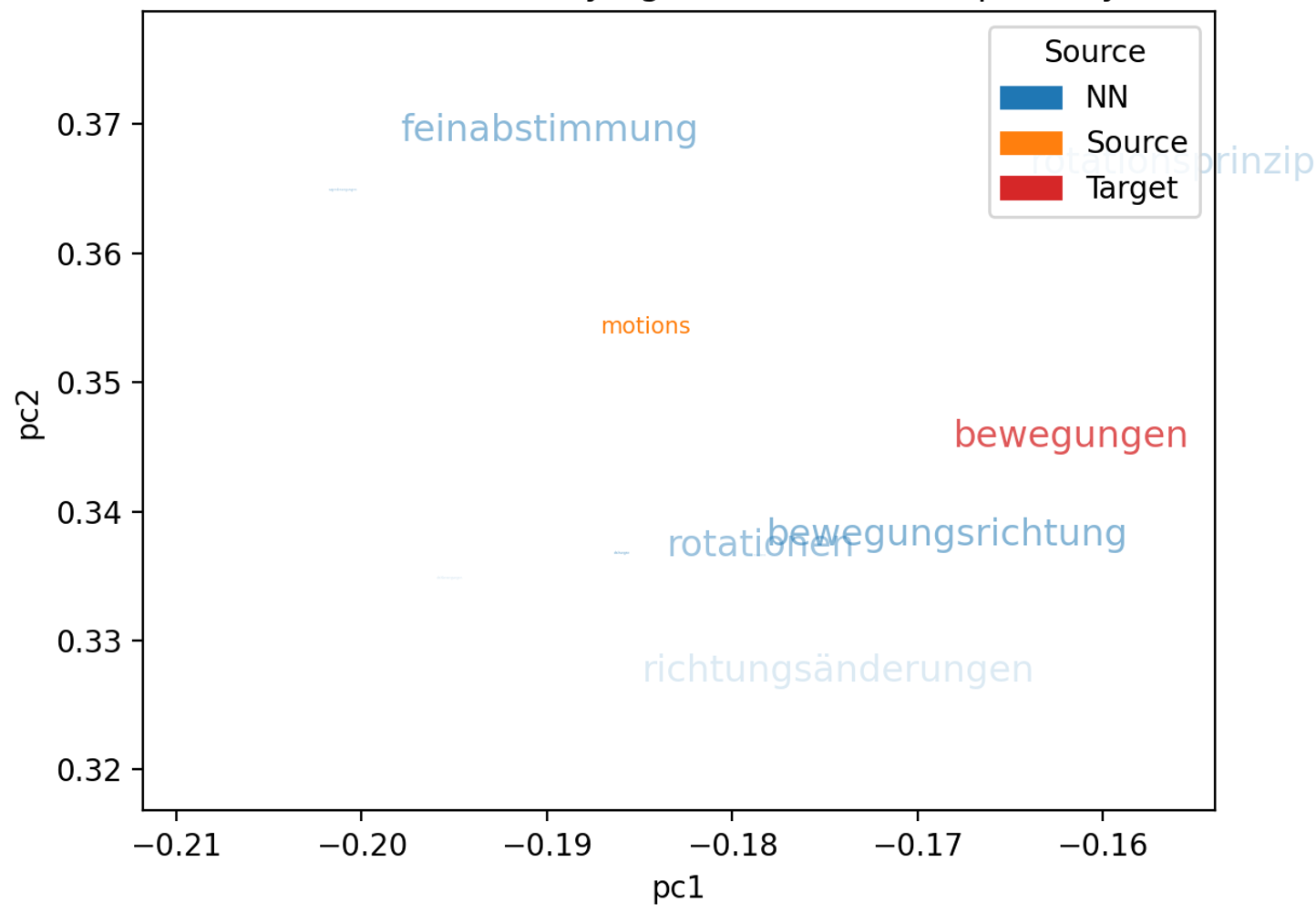}
\end{minipage}
\caption{Nearest Neighbours of "motions" with (left) and without (right) lexical features.}
\label{fig:example}
\end{figure*}

\appendix

\section{Qualitative Examples}
\label{sec:appendix}

In Table \ref{tab:examples}, we show select examples from the en-de test set where the LFBB model is able to successfully map source words to the correct target correspondences. The words predicted with the baseline C1 model are very close alternatives from the target languages which translate to rotations, monochromatic, and sword for the source words motions, coloured, and spear respectively. 

\begin{table}[h!]
    \small
    \centering
    \begin{tabular}{lccc}
    \toprule
         src &  motions&  coloured&  spear\\ 
         rank$_{src}$ &  15490&  8450&  13647\\ 
         pos$_{src}$ &  NOUN&  VERB&  NOUN\\ 
    \midrule
         pred$_{lfbb}$ &  bewegungen&  farbig&  speer\\ 
         rank$_{lfbb}$ &  5855&  19410&  15249\\
         pos$_{lfbb}$ &  NOUN&  ADV&  PROPN\\
    \midrule
         pred$_{c1}$ & rotationen& einfarbigen& schwert\\
         rank$_{c1}$ & 122792& 111085& 7149\\
         pos$_{c1}$ & NOUN& ADJ& VERB\\
    \bottomrule
    \end{tabular}
    \caption{Sample LFBB and C1 predictions (en-de)}
    \label{tab:examples}
\end{table}

The target words are much closer to the source words in relative frequency as shown by their ranks. The extra features help steer our model towards better predictions from amongst retrieved candidates that are very close in embedding space. We also plot the "motions" example in Figure \ref{fig:example}. The correct translation "bewegungen" is better highlighted after applying transparency and size re-scaling to indicate frequency difference and probability of part-of-speech match.



\end{document}